\documentclass[letterpaper]{article} 
\usepackage{aaai2026}
\usepackage{times}  
\usepackage{helvet}  
\usepackage{courier}  
\usepackage[hyphens]{url}  
\usepackage{graphicx} 
\usepackage{natbib}  
\usepackage{amsmath, amssymb}
\usepackage{caption} 
\usepackage{algorithm}
\usepackage{algorithmic}
\usepackage{booktabs}
\usepackage{multirow}

\usepackage{newfloat}
\usepackage{listings}
\DeclareCaptionStyle{ruled}{labelfont=normalfont,labelsep=colon,strut=off} 
\floatstyle{ruled}
\newfloat{listing}{tb}{lst}{}
\floatname{listing}{Listing}
\title{Revisiting the Data Sampling in Multimodal Post-training from a Difficulty-Distinguish View}
\author {
    Jianyu Qi\textsuperscript{\rm 1, 2*},
    Ding Zou\textsuperscript{\rm 2},
    Wenrui Yan\textsuperscript{\rm 2},
    Rui Ma\textsuperscript{\rm 2},
    Jiaxu Li\textsuperscript{\rm 1},
    Zhijie Zheng\textsuperscript{\rm 1},
    Zhiguo Yang\textsuperscript{\rm 2},
    Rongchang Zhao\textsuperscript{\rm 1, \dag}
}
\affiliations {
    \textsuperscript{\rm 1}School of Computer Science, Central South University, Changsha, Hunan, China\\
    \textsuperscript{\rm 2}Intelligent System Department, Zhongxing Telecom Equipment(ZTE), Changsha, Hunan, China\\
    qijianyu@csu.edu.cn, zoudinghust@gmail.com, ywraddlk@gmail.com, 214711069@csu.edu.cn,\\ lijiaxu@csu.edu.cn, zhengzhijie@csu.edu.cn, yang.zhiguo@zte.com.cn, zhaorc@csu.edu.cn
}

\begin{document}

\maketitle

\footnotetext[1]{Work done during internship at ZTE.} 
\footnotetext[2]{Corresponding author.} 

\begin{abstract}
Recent advances in Multimodal Large Language Models (MLLMs) have spurred significant progress in Chain-of-Thought (CoT) reasoning. Building on the success of Deepseek-R1, researchers extended multimodal reasoning to post-training paradigms based on reinforcement learning (RL), focusing predominantly on mathematical datasets. However, existing post-training paradigms tend to neglect two critical aspects: (1) The lack of quantifiable difficulty metrics capable of strategically screening samples for post-training optimization. (2) Suboptimal post-training paradigms that fail to jointly optimize perception and reasoning capabilities. To address this gap, we propose two novel difficulty-aware sampling strategies: Progressive Image Semantic Masking (PISM) quantifies sample hardness through systematic image degradation, while Cross-Modality Attention Balance (CMAB) assesses cross-modal interaction complexity via attention distribution analysis. Leveraging these metrics, we design a hierarchical training framework that incorporates both GRPO-only and SFT+GRPO hybrid training paradigms, and evaluate them across six benchmark datasets. Experiments demonstrate consistent superiority of GRPO applied to difficulty-stratified samples compared to conventional SFT+GRPO pipelines, indicating that strategic data sampling can obviate the need for supervised fine-tuning while improving model accuracy. Our code will be released at \url{https://github.com/qijianyu277/DifficultySampling}.

\end{abstract}

\section{Introduction}


Recent advancement of Multimodal Large Language Models (MLLMs) has witnessed a rapid development in Chain-of-Thought (CoT) reasoning \cite{wei2022chain, mitra2024compositional, kumar2025llm}, driven largely by innovative post-training techniques that align model behaviors with human reasoning patterns. These advancements are particularly significant as they enable MLLMs to handle complex tasks involving both visual and textual information more effectively. For instance, reinforcement learning frameworks like Group Relative Policy Optimization (GRPO) \cite{shao2024deepseekmath} have empowered MLLMs to autonomously discover reasoning paths through reward signals, enhancing their ability to perform intricate reasoning tasks.

Significant efforts have been devoted to enabling CoT reasoning in MLLMs
\cite{zhang2024improve,mitra2024compositional,zheng2023ddcot}. 
Previous methods
\cite{xu2024llava,zhang2023multimodal,guo2024mammoth,thawakar2025llamav} construct the datasets manually containing step-level reasoning processes and apply supervised
fine-tuning (SFT) to reformat MLLMs’ outputs, whose manually designed ``MLLs with formatted reasoning outputs" often results in ``Pseudo-CoT" reasoning \cite{huang2025vision, gao2024cantor, zheng2023ddcot}, lacking essential cognitive processes commonly observed in human thoughts.
Then with the success of Deepseek-R1 \cite{guo2025deepseek}, many researchers \cite{li2024superfiltering33,tong2024cambrian64,wang2022git68,luo2025ursa} have attempted to extend this success to multimodal reasoning,
where models process and reason on both visual and textual information. However, mainstream works \cite{liu2025noisyrollout, meng2025mm, chen2025unlocking} focus on performing RL with multimodal mathematical datasets, which improves the reasoning ability more in terms of textual modality but stresses less cross-modal ability.
More recently proposed methods \cite{ren2024pixellm, ma2025one, yu2025perception} focus on perception-enhanced RL training, via incorporating multimodal data (such as detection, grounding, etc.) into Reinforcement Learning with Verifiable Rewards (RLVR) and designing corresponding reward functions or models.

Despite effectiveness, current multimodal post-training methods commonly ignore two crucial questions:

\begin{itemize} 
    \item \textbf{How to identify multimodal data of distinct hardness?} 
    Recently proposed RL methods unanimously agreed that proper difficulty means a lot in RL training \cite{wang2025sota,wang2025dump,zhang2025r1}. But the core problem is that, multimodal data could not be divided into distinguished parts the same way as pure-text data (especically the math or code data), due to their multi-modal feature. It is clear that the text-modality difficulty could not be considered as the sample-difficulty for MLLM, and in many cases text-modality difficulty is not quantizable for MLLM (such as OCR and classify tasks). As a result, a suitable principle or definition for multi-modal hardness is supposed to consider cross-modal features, so as to correctly measure the hardness for MLLMs.
    \item \textbf{How to design effective post-training paradigm for MLLM?} 
    After distinguishing the hardness of various multi-modal data, it is necessary to design effective training pipline for MLLM. A straightfoward idea is to follow mainstream method to perform cold-start SFT with the hardest samples then RL with the medium difficulty. However, such a commonly-recognized paradigm is not always sufficient for MLLM post-training, owing to the multi-task features. Actually in post-training of MLLMs, multi-modal data could be roundly divided into two groups, i.e., Visual-Reasoning (Math, Science, Charting, and Puzzle, etc.) and Visual-Perception (Grounding, Counting, and OCR, etc.), it is necessary to design the optimal post-training scheme for each type of dataset.
    
\end{itemize}

To address the aforementioned issues, we define a hardness discrimination strategy for multimodal data from both intra-modal and cross-modal perspectives. We then uncover the distributional discrepancies between reasoning and perception data, and propose respective optimal post-training pipelines tailored to different types of datasets, aiming to enhance the reasoning and perception capabilities of multimodal learning models. Our main contributions can be summarized as follows:
\begin{itemize}
\item We propose the \textbf{Progressive Image Semantic Masking (PISM)}, a random masking strategy at the semantic level of images. By gradually increasing the masking ratio, we observe changes in the model's response state to determine the difficulty of samples.
\item We propose the \textbf{Cross-Modality Attention Balance (CMAB)}, an attention-based strategy that considers the attention scores of the response tokens generated by the model with respect to the text tokens and image tokens in the original input respectively. The degree of interaction between text and image tokens during response generation is measured by the ratio of their attention scores. Sample difficulty is then classified based on this interaction strength.
\item We design and verify the optimal training scheme for different types of datasets, rather than defaulting to supervised fine-tuning and reinforcement fine-tuning.
\end{itemize}

\section{Related Work}

\subsection{Multimodal Large Language Models Reasoning}

The rapid growth of Multimodal Large Language Models (MLLMs) \cite{chen2024image,ma2025one,su2025pixel} has endowed them with extensive knowledge and robust multitasking capabilities, enabling their application across complex and diverse domains, particularly for cross-modal tasks such as visual question answering and image captioning.
As the requirements for the capabilities of multimodal large models in real-world scenarios gradually increase, the ability of the thinking chain has also been gradually introduced into multimodal large models, such as Llava-COT \cite{xu2024llava}, leverage chainof-thought to enhance MLLMs reasoning capabilities. Other similar works employ MCTS-based methods to strengthen their reasoning abilities, such as Mulberry \cite{yao2024mulberry} which introduces collective knowledge to MCTS to search reasoning paths, thereby enhancing reasoning and reflection capabilities. 
However, such works which obtain CoT through supervised fine-tuning training often faces issues of weak generalization and high training costs due to the superfacial matching training, which encourages more researchers to design training paradigms that can adapt to the characteristics of large multimodal models to effectively guide complex reasoning processes.
Hence more recently proposed MLLMs focus on bringing RL-based reasoning into MLLMs, extend the DeepSeek-R1 training paradigm from large language models to MLLMs, such as VLM-R1 \cite{shen2025vlm}, Visual-RFT \cite{liu2025visual}, Vision-R1 \cite{huang2025vision}, and Perception-R1 \cite{yu2025perception}. 
\subsection{Reinforcement Learning for Post-Training}

With the popularity of DeepSeek-R1, several recent studies, including Open-Reasoner-Zero \cite{hu2025open}, SimpleRL-Zoo \cite{zeng2025simplerl}, and Logic-RL \cite{xie2025logic}, have explored directly RL post-training with GRPO, without any supplementary supervised fine-tuning stages. Additionally, complementary approaches such as VAPO \cite{yue2025vapo}, DAPO \cite{yu2025dapo}, and Dr.GRPO \cite{liu2025understanding} have sought to refine the GRPO framework by optimizing reward design and enhancing advantage estimation techniques, thus more effectively promoting deeper reasoning behaviors within language models. 
Meanwhile, recent analyses focus on the effect of data hardness on the RL training stage, such as \cite{huang2025vision} rank samples with human-pre-defined math difficulty, 
\cite{xiong2025minimalist} classify sample difficulty through rejection sampling, \cite{zhang2025grpo} arranges through reward score, and \cite{wang2025beyond} proposes to measure the hardness with the sentence entropy.
However, current multimodal post-training methods commonly ignoring the importance of data sampling in multi-modal data, neither ignore the data sampling operation nor filter with text-only principle, such as whether the problem is hard.
As a result, image-modal and mutli-interaction signals fail to be modeled, bring a suboptimal sampling strategy and further resulting worse model performance.
We hence propose to design a sampling strategy for multi-modal data from both intra-modal and inter-modal perspectives, which could involve cross-modal signals for multimodal post-training.

\section{Methodology}
Measurement of the difficulty of multimodal samples remains a fundamental challenge in post-training data curation. We introduce two complementary approaches that capture different aspects of sample complexity: \textbf{Progressive Image Semantic Masking
(PISM)} focusing on model sensitivity to visual perturbations, and \textbf{Cross-Modality Attention Balance
(CMAB)} examining the balance of cross-modal interactions during inference.

\begin{figure*}[t]
\centering
\includegraphics[width=1.90\columnwidth]{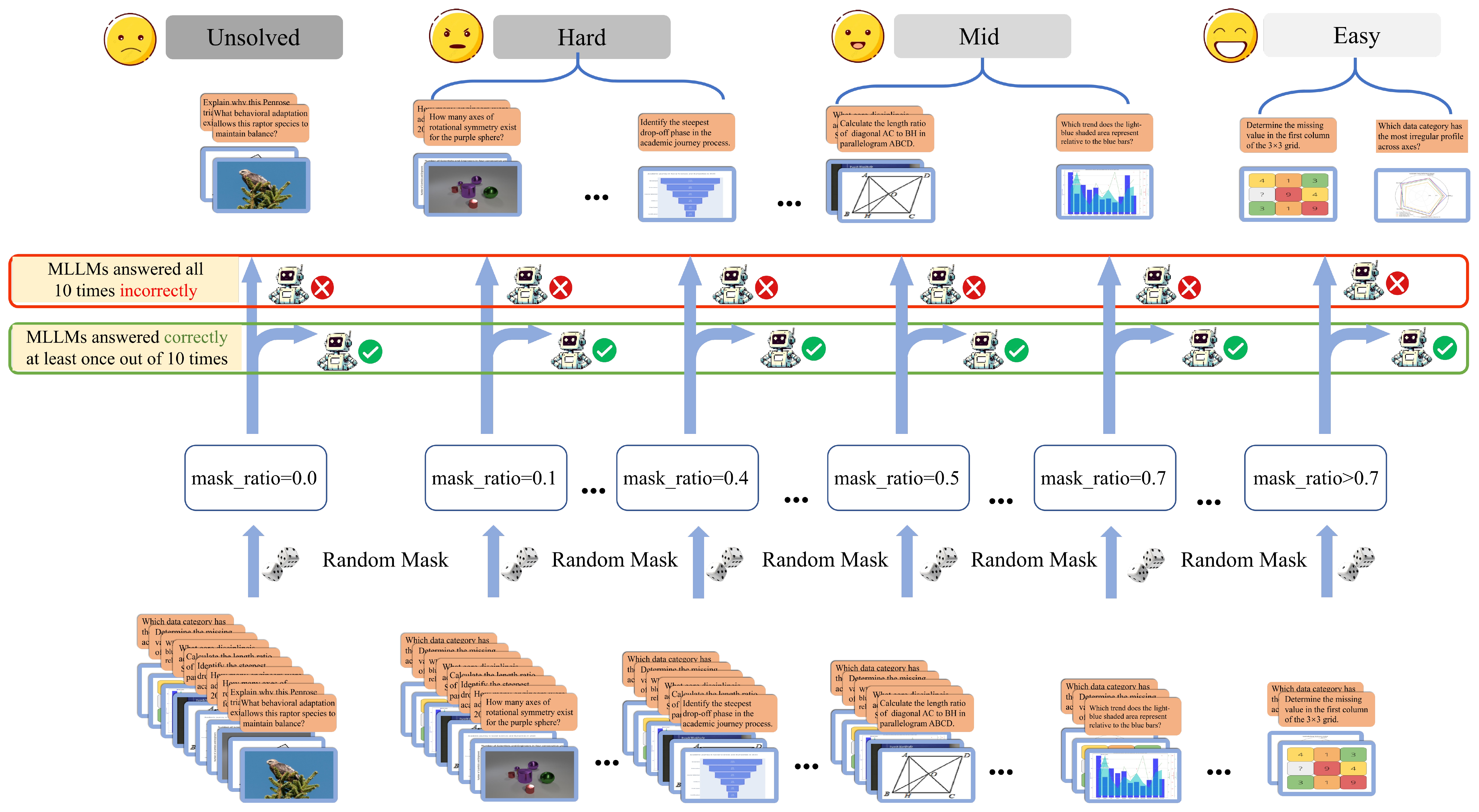}
\caption{Illustration of the \textbf{PISM} (Progressive Image Semantic Masking) method. We progressively mask different portions of the image, from no masking ($mask\_ratio=0.0$) to heavy masking ($mask\_ratio > 0.7$). Each masked image is created by randomly hiding a certain percentage of pixels. The process simulates varying levels of visual information loss. The model's performance is then evaluated on these masked images to understand how much it relies on visual details for accurate reasoning.}

\label{fig1}
\end{figure*}

\subsection{\textbf{PISM} for sensitivity-based Difficulty Assessment}

Intuitively, challenging multimodal samples should exhibit higher sensitivity to visual information loss—when critical visual content is obscured, the model's performance degrades more rapidly compared to easier samples. We operationalize this insight through a systematic masking strategy that gradually removes visual information while monitoring prediction stability.

\subsubsection{Mask-based Sensitivity Quantification}
Given an image-text pair $s = (I, Q)$, we systematically probe the model's visual dependence through controlled perturbation experiments. As shown in Figure \ref{fig1}, we define a series of masking ratios $\Lambda = \{ \lambda_i \mid \lambda_i = 0.0, 0.1, 0.2, \dots, 0.9 \}$, spanning from the original unmodified image ($\lambda = 0.0$) to heavily degraded versions where $90\%$ of pixels are occluded.

For each masking level $\lambda_i$, we apply the perturbation operation $M(\cdot, \lambda_i)$ that randomly selects and masks the specified proportion of pixels in the original image, yielding $I_{\lambda_i} = M(I, \lambda_i)$. This masking occurs directly in pixel space prior to any feature extraction, thereby simulating realistic scenarios of visual information loss or corruption that might occur in real-world applications.

We then evaluate model performance by feeding each perturbed sample $(I_{\lambda_i}, Q)$ to the multimodal model $\mathcal{M}$ and obtaining the prediction $A_{\lambda_i} = \mathcal{M}(I_{\lambda_i}, Q)$. The correctness of each prediction is assessed using a binary indicator:
\begin{equation}
    \delta_{\lambda_i} = 1[\mathcal{C}(A_{\lambda_i}, A_{\text{gt}})]
\end{equation}
where $\mathcal{C}$ evaluates whether the predicted answer matches the ground truth $A_{\text{gt}}$.

Given the stochastic nature of random masking, we repeat this evaluation process $K=10$ times with independent mask realizations for each masking ratio. The robust accuracy estimate is then computed as:
\begin{equation}
    P_c(\lambda_i) = \frac{1}{K} \sum_{k=1}^{K} \delta_{\lambda_i}^{(k)}
\end{equation}

The critical insight lies in identifying the failure threshold $\lambda_s^*$—the minimal masking ratio where performance drops below an acceptable level. Formally, we define:

\begin{equation}
    \lambda_s^* = \min \{ \lambda_i \in \Lambda \mid P_c(\lambda_i) < \tau \}
\end{equation}
where the threshold $\tau$ (we set $\tau = 0.1$)captures the transition point from reliable to unreliable predictions.

\begin{figure*}[t]
\centering
\includegraphics[width=1.95\columnwidth]{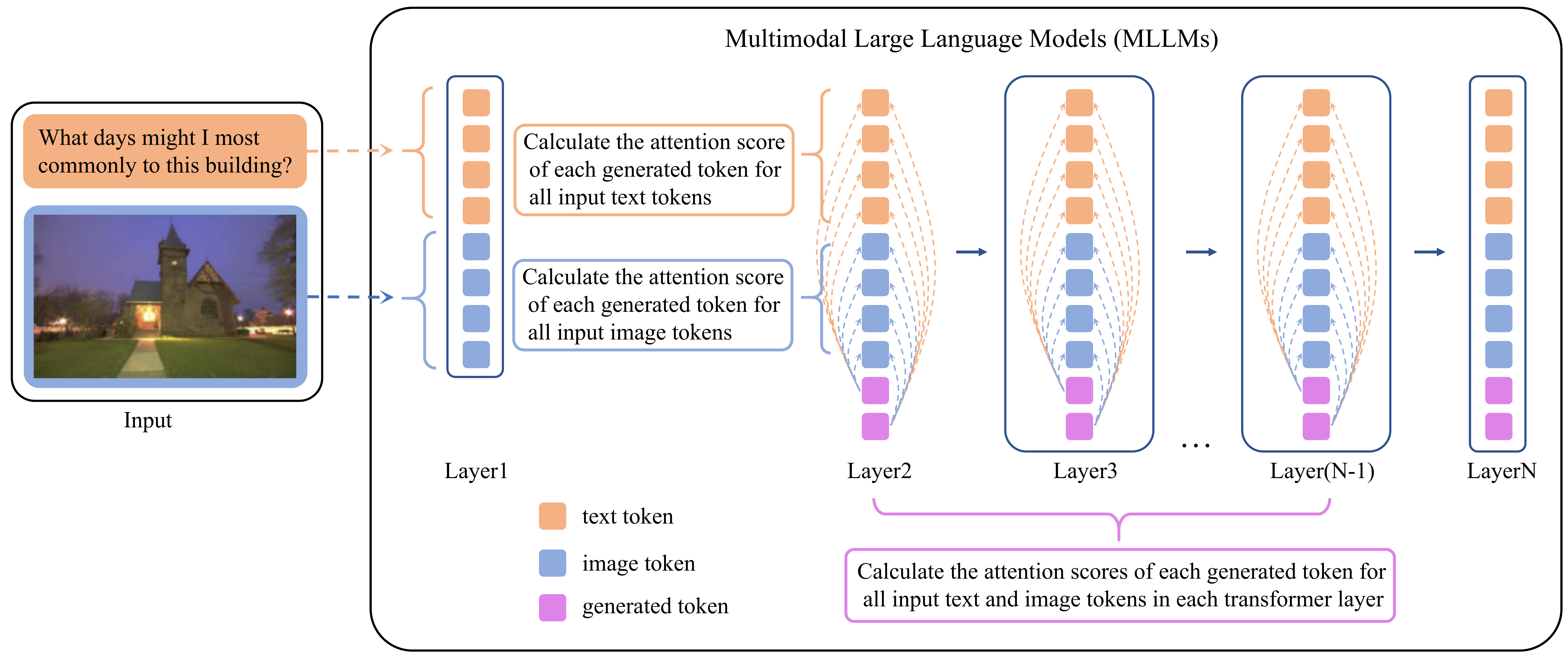}
\caption{Illustration of the \textbf{CMAB} (Cross-Modality Attention Balance) method. For each generated token, we calculate its average attention score over the input text tokens and image tokens across all transformer layers, and then average these scores across all generated tokens. $N$ represents the total number of layers of the transformer.}
\label{fig2}
\end{figure*}

\subsubsection{Sample Difficulty Classification via PISM}
The critical masking ratio $\lambda_s^*$ naturally partitions samples into distinct difficulty categories:
\begin{itemize}
    \item \textbf{Hard Samples:} $\lambda_s^* \leq \lambda_{\text{hard}}$ (we set $\lambda_{\text{hard}} = 0.4$). These samples exhibit fragility to minor visual perturbations, suggesting they require sophisticated visual understanding and tight multimodal coupling.

    \item \textbf{Medium Samples:} $\lambda_{\text{hard}} < \lambda_s^* < \lambda_{\text{easy}}$. These samples show moderate sensitivity to visual masking, indicating that while visual information contributes meaningfully to correct predictions, textual cues alone may still lead to partial or inconsistent performance. 
    
    \item \textbf{Easy Samples:} $\lambda_s^* \geq \lambda_{\text{easy}}$ (we set $\lambda_{\text{easy}} = 0.7$) or undefined (performance remains above $\tau$ across all masking levels). Such samples demonstrate robustness to visual degradation, indicating that textual information alone largely suffices for correct responses.
    
    \item \textbf{Unsolved Samples:} Samples with incorrect predictions on the original image ($P_c(0.0) < \tau$) are excluded from difficulty assessment, as their errors reflect model limitations rather than inherent sample complexity.
\end{itemize}

\subsection{CMAB for attention-based Difficulty Assessment}

While random masking methods based on the semantic layer of images can screen out difficult samples from the perspective of model sensitivity, it may not capture samples that require complex reasoning. We complement our approach by analyzing how the model allocates attention between different modalities during reasoning, which serves as a measure of the cognitive demands imposed by different sample types. A sample is considered difficult when the model allocates balanced attention to both text and images during question answering, as this indicates that information from both modalities is necessary to arrive at the correct answer.

\subsubsection{Attention-based Balance Quantification}
As shown in Figure.\ref{fig2}, for input sequences comprising image tokens $\mathbf{X}^{\text{img}} \in \mathbb{R}^{L_{\text{img}} \times d}$ and text tokens $\mathbf{X}^{\text{txt}} \in \mathbb{R}^{L_{\text{txt}} \times d}$, we analyze how the model distributes attention during the generation of each output token $y_t$ in the response sequence $\mathbf{Y} = \{y_1,\dots,y_T\}$.

During the generation of token $y_t$ at layer $l$, the cross-attention mechanism produces weights $\mathbf{A}^{(l, t)} \in \mathbb{R}^{1 \times (L_{\text{img}} + L_{\text{txt}})}$ that reveal the model's focus distribution across input tokens. We decompose this attention into modality-specific components by computing the total attention allocated to visual tokens as $S^{(l, t)}_{\text{img}} = \sum_{i=1}^{L_{\text{img}}} A^{(l, t)}_i$ and the attention to textual tokens as $S^{(l, t)}_{\text{txt}} = \sum_{i=L_{\text{img}}+1}^{L_{\text{img}} + L_{\text{txt}}} A^{(l, t)}_i$.

The ratio $\rho^{(l, t)} = S^{(l, t)}_{\text{img}} / S^{(l, t)}_{\text{txt}}$ captures the instantaneous attention balance at layer $l$ for token $t$. Values significantly above $1$ indicate visual dominance, while values below $1$ suggest textual preference. To obtain a stable estimate across the model's multiple layers, we compute the geometric mean:
\begin{equation}
    \rho_t = \exp \left( \frac{1}{L_{\text{layers}}-2} \sum_{l=2}^{L_{\text{layer}}-1} \log ( \rho^{(l, t)} + \epsilon ) \right)
\end{equation}
where $\epsilon \approx 10^{-8}$ prevents numerical instabilities when attention ratios approach zero. We exclude the first and last transformer layers when computing average attention scores, as they mainly handle input encoding and output decoding, with limited role in high-level semantic or cross-modal reasoning.

The sample-level attention balance emerges as the arithmetic mean across all generated tokens: $\bar{\rho} = \frac{1}{T} \sum_{t=1}^{T} \rho_t$. This metric provides a holistic view of how the model balances visual and textual information throughout the entire response generation process.

\subsubsection{Sample Difficulty Classification via CMAB}
Based on the attention score, we have the following difficulty classification:
\begin{itemize}
    \item \textbf{Easy Samples:} The attention balance ratio $\bar{\rho}$ serves as a window into the cognitive demands imposed by different sample types. For samples that the model answers correctly, extreme values of $\bar{\rho}$ typically indicate straightforward cases where one modality dominates the reasoning process. When $\bar{\rho} < 0.1$, the model relies heavily on textual information, suggesting that the visual content contributes minimally to the final answer. Conversely, when $\bar{\rho} > 1.9$, visual information takes precedence, often indicating questions that can be resolved through direct visual inspection without complex textual reasoning.
    \item \textbf{Medium Samples:} $0.1 \leq \bar{\rho} < 0.4$ or $1.6 < \bar{\rho} \leq 1.9$. These samples are primarily driven by one modality, but still require meaningful support from the other. They demand more than superficial integration, yet fall short of the balanced multimodal reasoning seen in hard samples.
    \item \textbf{Hard Samples:} The most intriguing cases emerge when $\bar{\rho}$ falls within a moderate range, approximately $[0.4, 1.6]$. This balanced attention allocation suggests that successful reasoning requires effective integration of both visual and textual information—neither modality alone provides sufficient context for arriving at the correct answer. Such samples typically involve complex spatial reasoning, visual-textual alignment, or nuanced interpretation that demands sophisticated cross-modal understanding.
    \item \textbf{Usolved Samples:} For samples that the model answers incorrectly, we categorize them as unsolved regardless of their attention patterns. While the attention distribution might provide insights into the model's reasoning process, the primary concern for difficulty assessment is whether the model can successfully leverage the available information to reach the correct conclusion.
\end{itemize}
The dual-perspective approach captures both the fragility of visual dependencies and the complexity of multimodal reasoning, providing a comprehensive understanding of sample difficulty.

\begin{table*}
\centering
\begin{tabular}{llllllllllll}
\midrule
Training paradigm     & MathVista & MMVet & OCRBench & HBench & MMMU & MMStar \\
\midrule        
GRPO-only(fullset)        & 53.400    & 41.697 & 76.200   & 67.403   & 0.440     & 0.607 \\
GRPO-only(unsolved)             & 67.200    & 50.183 & 78.700   & 69.295   & 0.537     & 0.629 \\
\midrule
SFT(mid)+GRPO(hard)        & 67.300 & 40.596 & 75.000 & 68.454 & 0.507 & 0.609 \\
SFT(mid)+GRPO(rand\_h)     & 67.700 & 38.578 & 74.800 & \textbf{69.085} & 0.504 & 0.606 \\
SFT(rand\_m)+GRPO(hard)    & 67.500 & 41.972 & 75.100 & 68.349 & 0.506 & 0.606 \\
SFT(rand\_m)+GRPO(rand\_h) & 66.600 & 42.248 & 74.700 & 68.980* & 0.509 & 0.609 \\
SFT(hard)+GRPO(mid)        & 67.300 & 39.312 & 74.200 & 67.613 & 0.502 & 0.608 \\
SFT(hard)+GRPO(rand\_m)    & 67.600 & 39.404 & 74.800 & 67.087 & 0.501 & 0.610 \\
SFT(rand\_h)+GRPO(mid)     & 67.200 & 42.661 & 74.700 & 68.559 & 0.504 & 0.598 \\
SFT(rand\_h)+GRPO(rand\_m) & 67.400 & 41.560 & 74.500 & 68.875 & 0.501 & 0.603 \\
\midrule
GRPO-only(random)          & 68.200* & \textbf{53.257} & 77.300* & 68.349 & 0.541* & 0.637* \\
GRPO-only(mid+hard)        & \textbf{68.300} & 48.257* & \textbf{77.800} & 68.770 & \textbf{0.547} & \textbf{0.639} \\
\bottomrule    
\end{tabular}
\caption{Comparison of training results using SFT+GRPO and GRPO-only on the visual reasoning dataset through PISM}
\label{tab:1}
\end{table*}

\section{Experiments}
\subsection{Experimental Setup}
Our empirical investigation employs the multimodal perception-reasoning benchmark established by \cite{ma2025one}. Experiments were executed on a computing cluster comprising five nodes equipped with NVIDIA A800-SXM4 GPUs (8×80GB memory per node) and two nodes with NVIDIA H20 GPUs (8×96GB memory per node). The implementation leverages PyTorch as the foundational computational framework, where supervised fine-tuning (SFT) was conducted using the LLaMA-Factory framework \cite{zheng2024llamafactory}, while GRPO training was implemented through the Swift framework \cite{zhao2025swift} for reward-constrained optimization. All methodologies were systematically evaluated on the Qwen2.5VL-7B foundation model \cite{bai2025qwen2}, ensuring controlled comparisons across experimental conditions.

\begin{table*}
\centering
\begin{tabular}{llllllllllll}
\midrule
Training paradigm          & MathVista & MMVet  & OCRBench & HBench & MMMU  & MMStar \\
\midrule  
GRPO-only(fullset)         & 70.000    & 51.147 & 77.200   & 68.034   & 0.557 & 0.615 \\
GRPO-only(unsolved)              & 69.200    & 52.385 & 77.700   & 68.875   & 0.541 & 0.609 \\
\midrule
SFT(mid)+GRPO(hard)        & 67.500    & 38.119 & 76.000   & 66.877   & 0.496 & 0.625 \\ 
SFT(mid)+GRPO(rand\_h)     & 67.400    & 38.624 & 75.800   & 66.141   & 0.498 & \textbf{0.628} \\
SFT(rand\_m)+GRPO(hard)    & 67.400    & 39.266 & 76.400   & 68.454   & 0.490 & 0.614 \\
SFT(rand\_m)+GRPO(rand\_h) & 66.500    & 44.771 & 75.600   & 69.085   & 0.498 & 0.605 \\
SFT(hard)+GRPO(mid)        & 67.900*   & 42.844 & 76.500   & 68.033   & 0.512 & 0.625 \\
SFT(hard)+GRPO(rand\_m)    & 67.500    & 41.101 & 76.000   & 68.454   & 0.512 & 0.610 \\
SFT(rand\_h)+GRPO(mid)     & 65.600    & 45.459 & 75.000   & 68.770*  & 0.503 & 0.612 \\
SFT(rand\_h)+GRPO(rand\_m) & 67.700    & 37.982 & 75.700   & 68.665   & 0.488 & 0.613 \\
\midrule
GRPO-only(random)          & \textbf{68.100}    & 49.908* & 77.300*   & 68.559 & \textbf{0.553} & 0.627* \\
GRPO-only(mid+hard)        & 67.600    & \textbf{52.477} & \textbf{77.600}   & \textbf{69.716} & 0.544* & 0.625 \\
\hline
\end{tabular}
\caption{Comparison of training results using SFT+GRPO and GRPO-only on the visual perception dataset through PISM}
\label{tab:2}
\end{table*}

\begin{table*}
\centering
\begin{tabular}{lllllllll}
\midrule
Training paradigm  & MathVista & MMVet  & OCRBench & HBench   & MMMU      & MMStar \\
\midrule  
GRPO-only(fullset)        & 53.400    & 41.697 & 76.200   & 67.403   & 0.440     & 0.607 \\
GRPO-only(unsolved)       & 69.340    & 54.450 & 78.300   & 68.244   & 0.547     & 0.636 \\
\midrule 
SFT(mid)+GRPO(hard)       & 67.200    & 33.486 & 74.300   & 67.298   & 0.499     & 0.627 \\ 
SFT(mid)+GRPO(rand\_h)    & 67.700    & 33.991 & 74.300   & 65.615   & 0.508     & 0.621 \\
SFT(rand\_m)+GRPO(hard)   & 67.800    & 34.541 & 72.300   & 67.718   & 0.502     & 0.625 \\
SFT(rand\_m)+GRPO(rand\_h)& 67.600    & 33.394 & 73.500   & 64.984   & 0.498     & 0.622 \\
SFT(hard)+GRPO(mid)       & 67.400    & 34.037 & 75.200   & 68.244*  & 0.501     & 0.618 \\
SFT(hard)+GRPO(rand\_m)   & 66.900    & 36.422 & 74.400   & 68.034   & 0.499     & 0.619 \\
SFT(rand\_h)+GRPO(mid)    & 67.300    & 34.541 & 75.300   & 68.139   & 0.500     & 0.617 \\
SFT(rand\_h)+GRPO(rand\_m)& 67.700    & 34.266 & 74.200   & 67.087   & 0.494     & 0.619 \\
\midrule 
GRPO-only(random)         & 68.200*    & 43.624* & \textbf{77.300}   & 69.085    & \textbf{0.556}     & \textbf{0.642} \\
GRPO-only(mid+hard)       & \textbf{69.000} & \textbf{48.578} & 77.100*  & \textbf{69.085}  & 0.542*  & 0.628* \\
\hline
\end{tabular}
\caption{Comparison of training results using SFT+GRPO and GRPO-only on the visual reasoning dataset through CMAB}
\label{tab:3}
\end{table*}

\begin{table*}
\centering
\begin{tabular}{lllllllll}
\midrule
Training paradigm  & MathVista & MMVet  & OCRBench & HBench   & MMMU      & MMStar \\
\midrule  
GRPO-only(fullset)        & 70.000    & 51.147 & 77.200   & 68.034   & 0.557     & 0.615 \\
GRPO-only(unsolved)       & 68.700    & 54.541 & 77.700   & 69.085   & 0.536     & 0.615 \\
\midrule 
SFT(mid)+GRPO(hard)       & 66.800    & 41.239 & 75.100   & 68.244   & 0.503     & 0.627* \\ 
SFT(mid)+GRPO(rand\_h)    & 66.500    & 42.431 & 75.200   & 67.823   & 0.499     & 0.626 \\
SFT(rand\_m)+GRPO(hard)   & 67.800    & 36.514 & 75.100   & 68.875   & 0.499     & 0.625 \\
SFT(rand\_m)+GRPO(rand\_h)& 67.500    & 42.798 & 75.000   & 68.769   & 0.496     & 0.623 \\
SFT(hard)+GRPO(mid)       & 67.400    & 34.037 & 75.200   & 68.244   & 0.501     & 0.618 \\
SFT(hard)+GRPO(rand\_m)   & 67.900    & 48.945 & 75.900   & 67.718   & 0.538     & 0.609 \\
SFT(rand\_h)+GRPO(mid)    & 68.100*   & 49.500 & 76.500   & 68.454   & 0.534     & 0.607 \\
SFT(rand\_h)+GRPO(rand\_m)& 67.600    & 50.321* & \textbf{77.500}   & 68.980*   & 0.526     & 0.610 \\
\midrule 
GRPO-only(random)         & 67.700    & 45.550 & 76.900*   & \textbf{69.401}   & 0.545*     & 0.625 \\
GRPO-only(mid+hard)       & \textbf{68.300}    & \textbf{50.367} & 76.800  & 68.244  & \textbf{0.550}  & \textbf{0.629} \\ 
\hline
\end{tabular}
\caption{Comparison of training results using SFT+GRPO and GRPO-only on the visual perception dataset through CMAB}
\label{tab:4}
\end{table*}


\subsection{Evaluation Framework and Benchmarks}
Our evaluation encompasses general visual question answering through MMVet \cite{yu2023mm}, which test fundamental visual comprehension and reasoning abilities. Mathematical and chart interpretation capabilities are assessed using MathVistar \cite{lu2023mathvista} and MMMU \cite{yue2024mmmu}, both requiring sophisticated numerical reasoning and visual analysis. For multimodal knowledge integration and complex reasoning, we employ MMStar \cite{chen2024right}. Document understanding and optical character recognition are evaluated through OCRBench \cite{liu2024ocrbench}, while hallucination detection capabilities are measured via HallusionBench \cite{guan2024hallusionbench}. All benchmarks are evaluated within the OpenCompass framework \cite{buitrago2019open}, using GPT-4o-mini \cite{hurst2024gpt} as the unified judge model for consistent and reliable scoring.

\subsection{Training Setup: GRPO-Only vs. SFT+GRPO}
We use \textbf{PISM} and \textbf{CMAB} to classify samples in the perception and reasoning datasets into difficulty levels, respectively. The resulting sample distributions and associated training strategies are detailed in \textit{Appendix Tables A–D}.
Based on these classifications, we compare two post-training paradigms:
\textbf{(1) GRPO-only}, which applies Group Relative Policy Optimization directly to the full dataset; and
\textbf{(2) SFT+GRPO}, which first performs supervised fine-tuning (SFT) on a curated subset before applying GRPO.
Our goal is to evaluate whether pure reinforcement learning fine-tuning or a hybrid approach yields better performance on perception and reasoning tasks. For the SFT+GRPO paradigm, we further investigate the impact of training order—specifically, how the sequencing of medium- and high-difficulty samples during SFT influences final performance—enabling a more refined training paradigm.

\subsection{Results and Analysis}
The final experimental results are reported in Table \ref{tab:1} and Table \ref{tab:2}, which present performance metrics (HBench stands for HallusionBench) for visual perception and visual reasoning datasets, respectively. The \textbf{bold} numbers represent the best results, and the \textbf{*} represents the suboptimal results. Here, ``mid" denotes medium samples, ``rand\_m" refers to a random dataset of the same size as the medium sample, and ``rand\_h" refers to a random dataset of the same size as the hard samples. It is worth noting that due to the large amount of data in the full dataset and unsolved data, their GRPO-only results are only used as a reference and are not included in the comparison of results with other training strategies.

\subsubsection{Efficacy of Difficulty-Aware Sampling Strategies}
The experimental results consistently validate the superiority of our proposed difficulty-aware sampling strategies (PISM and CMAB) across both visual perception and reasoning tasks. As demonstrated in Tables \ref{tab:1}--\ref{tab:4}, models trained on difficulty-stratified samples (mid+hard) using GRPO-only paradigm outperform those trained with conventional SFT+GRPO pipelines, highlighting the critical role of strategic data selection in multimodal post-training. \textbf{PISM} exhibits particular strength in tasks requiring robust visual perception. On OCRBench, which evaluates optical character recognition capabilities, the GRPO-only(mid+hard) configuration achieves a score of 77.800 (Table \ref{tab:1}), surpassing all SFT+GRPO variants by at least 1.3 points. This performance gain can be attributed to PISM's ability to identify samples where visual information is irreplaceable—by systematically masking image pixels and measuring performance degradation, PISM effectively isolates samples that demand precise visual understanding. The significant drop in performance when using random samples (GRPO-only(random): 77.300) confirms that indiscriminate data inclusion dilutes training efficiency by introducing redundant samples where textual cues suffice. \textbf{CMAB} shows greater efficacy in complex reasoning tasks that require tight integration of visual and textual information. On MathVista, which assesses mathematical reasoning in visual contexts, CMAB-stratified mid+hard samples yield a GRPO-only score of 69.000 (Table \ref{tab:3}), outperforming PISM-based training (68.300) and all SFT+GRPO configurations. This advantage stems from CMAB's unique capability to quantify cross-modal interaction complexity—by analyzing attention distribution between image and text tokens, CMAB identifies samples requiring balanced multimodal processing ($\rho \in [0.4, 1.6]$). The consistent outperformance on MMStar (0.639 vs. 0.625 in SFT+GRPO) further validates that attention balance is a robust indicator of reasoning difficulty.
Notably, both strategies demonstrate complementary strengths: PISM excels in perception-heavy tasks (OCRBench, MMVet) by focusing on visual sensitivity, while CMAB dominates reasoning tasks (MathVista, MMMU) through attention analysis. This complementarity confirms that multimodal difficulty assessment requires a dual perspective encompassing both intra-modal sensitivity and inter-modal interaction.

\subsubsection{Superiority of GRPO-Only Paradigm}
Our experiments reveal a counterintuitive finding: GRPO-only training on difficulty-stratified samples consistently outperforms the widely adopted SFT+GRPO pipeline across all evaluated benchmarks. This challenges the prevailing assumption that supervised fine-tuning is a necessary prerequisite for effective reinforcement learning in multimodal systems.

The key advantage of GRPO-only lies in its ability to avoid ``Pseudo-CoT" reasoning patterns induced by SFT. As shown in HallusionBench results (Table \ref{tab:2}), GRPO-only(mid+hard) achieves a score of 69.716, exceeding SFT+GRPO variants which top out at 68.980. This indicates that SFT's reliance on manually designed reasoning templates may encourage surface pattern matching rather than genuine logical reasoning, increasing hallucination risk. In contrast, GRPO's reward-driven optimization directly reinforces correct reasoning paths without constraining the model to artificial templates.

Another critical observation is the performance gap between difficulty-stratified and full-dataset GRPO training. On MathVista, GRPO-only(mid+hard) using PISM achieves 68.300, substantially outperforming GRPO-only(fullset) at 53.400 (Table \ref{tab:1}). This discrepancy highlights the inefficiency of training on unfiltered data, where easy samples and unsolved cases dilute the signal from informative mid+hard samples. The similar trend observed across other benchmarks (MMMU: 0.547 vs. 0.440; MMStar: 0.639 vs. 0.607) confirms that strategic sample selection is more impactful than simply increasing data volume.

These results highlight two key insights: (1) Reinforcement learning, when guided by difficulty-aware sample selection, can effectively learn perception and reasoning without prior supervised fine-tuning. (2) Data quality—particularly the strategic inclusion of medium and hard samples—matters more than quantity in driving multimodal performance. This approach not only simplifies the training pipeline by removing the SFT stage, but also improves model robustness by focusing on samples that truly challenge multimodal reasoning.

\section{Conclusion}
In this work, we propose two difficulty-aware sample selection methods—Progressive Image Semantic Masking (PISM) and Cross-Modal Attention Balancing (CMAB)—to enable fine-grained difficulty classification for multimodal data. These metrics assess visual sensitivity and cross-modal alignment, guiding more strategic post-training. Experiments show that GRPO-only training consistently outperforms SFT+GRPO on both visual perception and reasoning tasks, especially on medium and hard samples. This indicates that, with well-chosen samples, reinforcement learning can effectively learn visual and logical reasoning without supervised fine-tuning. The success of the GRPO-only paradigm challenges the assumption that SFT is necessary for stable alignment, suggesting that direct policy optimization on informative samples better preserves reasoning capabilities and avoids overfitting. Our results emphasize intelligent data use over complex training pipelines, pointing to a simpler, more effective path toward multimodal alignment.

\bibliography{aaai2026}


\begin{table*}[t]
\centering
\renewcommand{\thetable}{A}
\label{tab:PISM_classification}
\begin{tabular}{@{}lclcccc@{}}
\toprule
\multirow{2}{*}{Task Type} & \multirow{2}{*}{Metric} & \multirow{2}{*}{Overall Data} & \multicolumn{4}{c}{Difficulty Classification} \\ 
\cmidrule{4-7}
& & & Easy & Medium & Hard & Unsolved \\ 
\midrule
\multirow{3}{*}{Visual Perception} 
& Mask Ratio & \multirow{3}{*}{20,633} & $(0.7,1)$ & $(0.4,0.7]$ & $(0,0.4]$ & $0$ \\
& Difficulty Level & & easy & mid & hard & unsolved \\
& Data Volume & & 7,827 & 4,872 & 1,454 & 6,480 \\ 
\midrule
\multirow{3}{*}{Visual Reasoning} 
& Mask Ratio & \multirow{3}{*}{27,133} & $(0.7,1)$ & $(0.4,0.7]$ & $(0,0.4]$ & $0$ \\
& Difficulty Level & & easy & mid & hard & unsolved \\
& Data Volume & & 5,048 & 1,061 & 1,618 & 19,406 \\ 
\bottomrule
\end{tabular}
\caption{Sample distribution and difficulty classification using PISM (Progressive Image Semantic Masking) method.}
\end{table*}

\begin{table*}[t]
\centering
\renewcommand{\thetable}{B}
\label{tab:PISM_training}
\begin{tabular}{@{}llll@{}}
\toprule
Task Type & Strategy & \multicolumn{2}{c}{Data Subsets and Combinations} \\
\cmidrule(lr){3-4}
& & Combination A & Combination B \\
\midrule
\multirow{5}{*}{Visual Perception} 
& GRPO-only & \multicolumn{2}{l}{mid+hard (6.3k), random (6.3k), unsolved (6k), fullset (20.6k)} \\
& SFT+GRPO-1 & mid-4.9k (SFT) + hard-1.4k (GRPO) & hard-1.4k (SFT) + mid-4.9k (GRPO) \\
& SFT+GRPO-2 & random-4.9k (SFT) + hard-1.4k (GRPO) & random-1.4k (SFT) + mid-4.9k (GRPO) \\
& SFT+GRPO-3 & mid-4.9k (SFT) + random-1.4k (GRPO) & hard-1.4k (SFT) + random-4.9k (GRPO) \\
& SFT+GRPO-4 & random-1.4k (SFT) + random-4.9k (GRPO) & random-4.9k (SFT) + random-1.4k (GRPO) \\
\midrule
\multirow{5}{*}{Visual Reasoning} 
& GRPO-only & \multicolumn{2}{l}{mid+hard (2.6k), random (2.6k), unsolved (19k), fullset (27k)} \\
& SFT+GRPO-1 & mid-1k (SFT) + hard-1.6k (GRPO) & hard-1.6k (SFT) + mid-1k (GRPO) \\
& SFT+GRPO-2 & random-1k (SFT) + hard-1.6k (GRPO) & random-1.6k (SFT) + mid-1k (GRPO) \\
& SFT+GRPO-3 & mid-1k (SFT) + random-1.6k (GRPO) & hard-1.6k (SFT) + random-1k (GRPO) \\
& SFT+GRPO-4 & random-1k (SFT) + random-1.6k (GRPO) & random-1k (SFT) + random-1k (GRPO) \\
\bottomrule
\end{tabular}
\caption{Training configurations for PISM-based experiments using GRPO-only and SFT+GRPO strategies.}
\end{table*}

\begin{table*}[t]
\centering
\renewcommand{\thetable}{C}
\label{tab:CMAB_classification}
\begin{tabular}{@{}lclcccc@{}}
\toprule
\multirow{2}{*}{Task Type} & \multirow{2}{*}{Metric} & \multirow{2}{*}{Overall Data} & \multicolumn{4}{c}{Difficulty Classification} \\ 
\cmidrule{4-7}
& & & Easy & Medium & Hard & Unsolved \\ 
\midrule
\multirow{2}{*}{Visual Perception} 
& Range of $\bar{\rho}$ & \multirow{2}{*}{20633} & $(0,0.1) \cup (1.9,+\infty)$ & $[0.1,0.4) \cup (1.6,1.9]$ & $[0.4,1.6]$ & -- \\
& Data Volume & & 6753 & 6029 & 1001 & 6850 \\ 
\midrule
\multirow{2}{*}{Visual Reasoning} 
& Range of $\bar{\rho}$ & \multirow{2}{*}{27133} & $(0,0.1) \cup (1.9,+\infty)$ & $[0.1,0.4) \cup (1.6,1.9]$ & $[0.4,1.6]$ & -- \\
& Data Volume & & 2170 & 3604 & 2166 & 19193 \\ 
\bottomrule
\end{tabular}
\caption{Sample distribution and difficulty classification using the CMAB (Cross-Modality Attention Balance) method.}
\end{table*}

\begin{table*}[t]
\centering
\renewcommand{\thetable}{D}
\label{tab:training_config}
\begin{tabular}{@{}llll@{}}
\toprule
Task Type & Strategy & \multicolumn{2}{c}{Data Subsets and Sample Sizes} \\
\cmidrule(lr){3-4}
& & Combination A & Combination B \\
\midrule
\multirow{5}{*}{Visual Perception} 
& GRPO-only & \multicolumn{2}{l}{mid+hard (7k), random (7k), unsolved (6.8k), fullset (20.6k)} \\
& SFT+GRPO-1 & mid-6k (SFT) + hard-1k (GRPO) & hard-1k (SFT) + mid-6k (GRPO) \\
& SFT+GRPO-2 & random-6k (SFT) + hard-1k (GRPO) & random-1k (SFT) + mid-6k (GRPO) \\
& SFT+GRPO-3 & mid-6k (SFT) + random-1k (GRPO) & hard-1k (SFT) + random-6k (GRPO) \\
& SFT+GRPO-4 & random-6k (SFT) + random-1k (GRPO) & random-1k (SFT) + random-6k (GRPO) \\
\midrule
\multirow{5}{*}{Visual Reasoning} 
& GRPO-only & \multicolumn{2}{l}{mid+hard (5.7k), random (5.7k), unsolved (19k), fullset (27k)} \\
& SFT+GRPO-1 & mid-3.6k (SFT) + hard-2.1k (GRPO) & hard-2.1k (SFT) + mid-3.6k (GRPO) \\
& SFT+GRPO-2 & random-3.6k (SFT) + hard-2.1k (GRPO) & random-2.1k (SFT) + mid-3.6k (GRPO) \\
& SFT+GRPO-3 & mid-3.6k (SFT) + random-2.1k (GRPO) & hard-2.1k (SFT) + random-3.6k (GRPO) \\
& SFT+GRPO-4 & random-3.6k (SFT) + random-2.1k (GRPO) & random-2.1k (SFT) + random-3.6k (GRPO) \\
\bottomrule
\end{tabular}
\caption{Training configurations for CMAB-based experiments using GRPO-only and SFT+GRPO strategies.}
\end{table*}

\end{document}